\documentclass[10pt,twocolumn,letterpaper]{article}
\usepackage[pagenumbers]{cvpr} 

\usepackage{graphicx}
\usepackage{amsmath}
\usepackage{amssymb}
\usepackage{booktabs}
\usepackage{pifont}
\newcommand{\cmark}{\ding{51}}%
\newcommand{\xmark}{\ding{55}}%
\usepackage[accsupp]{axessibility}  
\usepackage{multirow}
\usepackage[table,xcdraw,dvipsnames]{xcolor}

\definecolor{captioning_col}{HTML}{189f7b}
\definecolor{instr_col}{HTML}{593da7}

\newcommand{\fparagraph}[1]{\paragraph{#1}}

\usepackage[pagebackref,breaklinks,colorlinks]{hyperref}

\usepackage[capitalize]{cleveref}
\crefname{section}{Sec.}{Secs.}
\Crefname{section}{Section}{Sections}
\Crefname{table}{Table}{Tables}
\crefname{table}{Tab.}{Tabs.}

\begin{document}

\title{Learning Action Changes by Measuring Verb-Adverb Textual Relationships}

\author{Davide Moltisanti, Frank Keller, Hakan Bilen, Laura Sevilla-Lara\\
The University of Edinburgh, United Kingdom\\
{\tt\small \{davide.moltisanti, frank.keller, h.bilen, l.sevilla\}@ed.ac.uk}
}
\maketitle

\begin{abstract}
The goal of this work is to understand the way actions are performed 
in videos. That is, given a video, we aim to predict an adverb indicating a modification applied to the action 
(e.g. cut ``finely''). 
We cast this problem as a regression task. 
We measure textual relationships between verbs and adverbs to generate a regression target representing the action change we aim to learn.
We test our approach on a range of datasets and achieve state-of-the-art results on both adverb prediction and antonym classification.
Furthermore, we outperform previous work when we lift two commonly assumed 
conditions: 
the availability of action labels during testing and the pairing of adverbs as antonyms. 

Existing datasets for adverb recognition 
are either noisy, which makes learning difficult, or contain actions whose
appearance is not influenced by adverbs, which makes evaluation less reliable.
To address this, we collect a new high quality dataset: Adverbs in Recipes (AIR). We focus on instructional recipes videos, curating a set of actions that exhibit meaningful visual changes when performed differently. Videos in AIR are more tightly trimmed and 
were manually reviewed by multiple annotators to ensure high labelling quality. 
Results show that models learn better from AIR given its cleaner videos. At the same time, adverb prediction on AIR is challenging, demonstrating that there is considerable room for improvement.

\end{abstract}

\vspace{-10pt}
\section{Introduction}
\label{sec:intro}

\begin{figure}
    \centering
    \includegraphics[width=0.9\columnwidth]{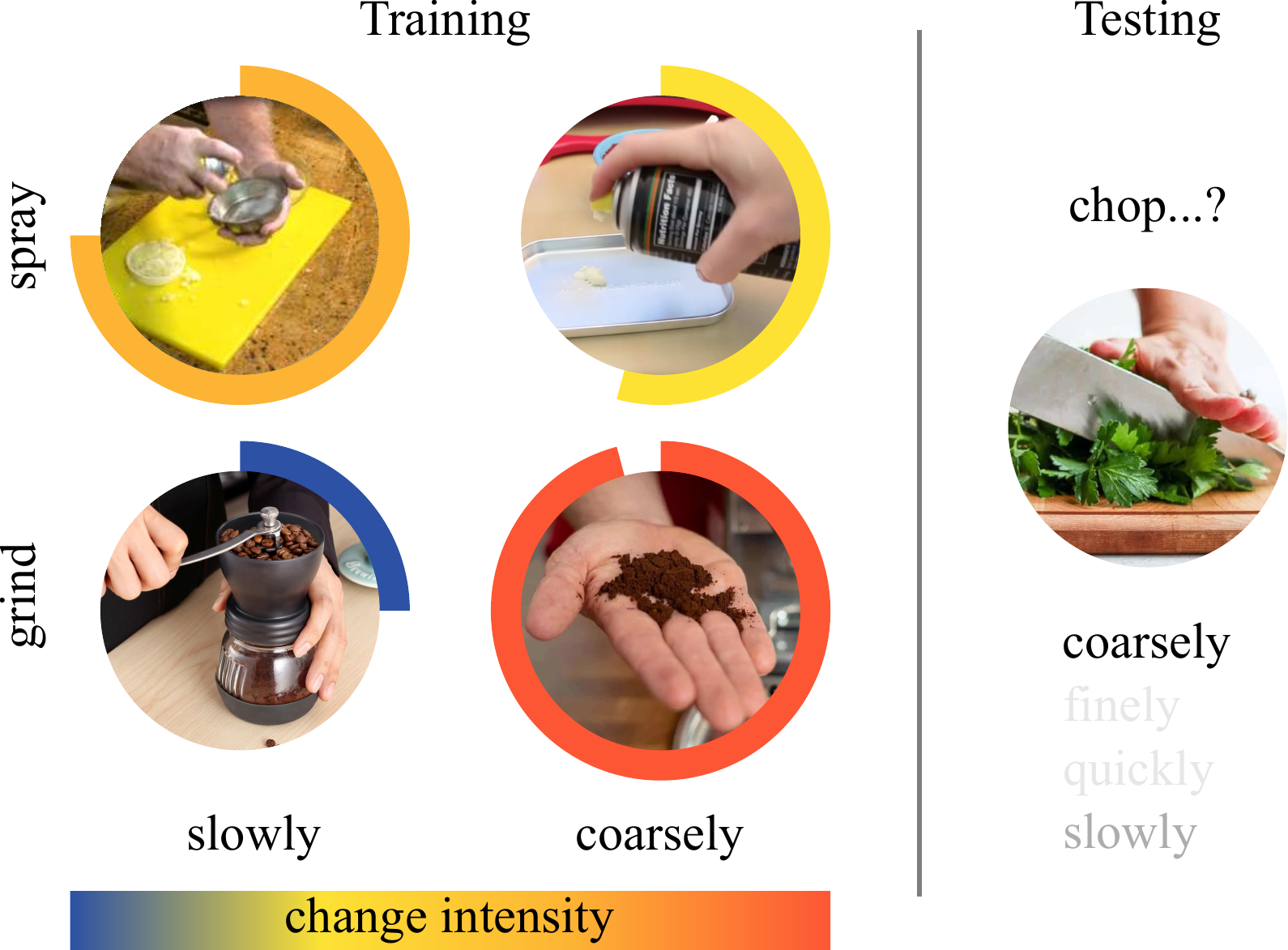}
    \caption{We aim to predict the way an action is performed in a video (right). Actions and their outcomes change in different ways when modified by adverbs, e.g. grinding coffee \textit{slowly} does not have the same effect as grinding it \textit{coarsely} (left). We learn to predict adverbs by recognising action changes in the video. We define action changes measuring verb-adverb textual relationships.}
    \label{fig:teaser}    
\end{figure}

Learning  
how an action is performed in a video is an important step towards expanding our video understanding beyond 
action recognition.  
This task has been referred to as ``adverb recognition'', and 
has potential useful applications in robotics and retrieval. Consider a scenario where a robot 
handles fragile objects. In such case we would like to tell the robot to grasp objects \textit{gently} 
to prevent it from breaking things. Similarly, learning how actions are performed enables more sophisticated queries for 
retrieval. Imagine 
learning a new recipe where a crucial step is stirring a mixture \textit{vigorously}. 
It would be useful to find good examples showing how to whip a mixture in such a manner. 

Adverb recognition is a challenging problem. Firstly, a single adverb can modify multiple actions in different ways. For example, 
to \textit{grind} coffee \textit{coarsely} (see Figure~\ref{fig:teaser}) we need to set a certain granularity in a grinder. Conversely, to \textit{spray} something \textit{coarsely} we would just use a spraying can more sparingly. 
Secondly, compared to actions or objects, adverbs are much harder to identify. 
Though somewhat subjective~\cite{moltisanti2017trespassing,alwassel2018diagnosing}, the temporal boundaries of an action can be easily determined. The spatial extent of an object is also easy to recognise for a human. 
Adverbs are instead more abstract: 
we can see their effect, but we struggle to pinpoint 
the spatio-temporal location where the adverb is appearing. 
Consider for example the act of spraying something \textit{slowly} 
in Figure~\ref{fig:teaser}. 
We could say that \textit{slowly} ``is located'' on the 
hand motions. At the same time, spraying something slowly could implicate a more even coating of a surface.
That is, the outcome of an action and the end-state of the involved objects can also change according to the adverb. 
This makes it hard to label adverbs with visual annotations (e.g. bounding boxes), 
which makes the problem more challenging. 

In fact, previous approaches~\cite{doughty2020action, doughty2022you} learn adverbs as action changes in a weakly supervised manner. The state-of-the-art approach~\cite{doughty2020action} treats adverbs as learnable parameters that modify actions. Specifically, 
the action change is learnt during training by contrasting antonyms, i.e. opposite adverbs.
We show that learning adverbs in such manner 
can be difficult and can limit the ability of the model to generalise.
We thus propose to define action changes by measuring distances in a text embedding space, and aim to learn such change from the video through regression. We show that 
our approach achieves new state-of-the-art results through extensive experiments. We also lift two major assumptions made in previous work~\cite{doughty2020action, doughty2022you}: the availability of action labels during testing and the pairing of opposite adverbs as antonyms. 
Our method achieves stronger performance especially when the above assumptions are relaxed. 

Besides being challenging, adverb recognition is also an under-explored domain and only few datasets are available.
Doughty and Snoek~\cite{doughty2022you} addressed the problem of scarce annotations, proposing a semi-supervised method that assigns pseudo-labels 
to boost recognition performance.  
Three datasets sourced from captioning benchmarks were also collected in~\cite{doughty2022you}. 
These datasets offer a large number of clips, actions and adverbs. However, 
adverbs in these datasets appear to be 
descriptive rather than action modifiers, i.e. actions do not display a significant change when modified by the adverb. 
This is an issue when adverbs are modelled as action changes as in this work. 
We thus introduce a new dataset focusing on instructional videos where actions change considerably depending on the way they are carried out. We focus on cooking videos since in this domain action changes are prominent. Our \textbf{Adverbs in Recipes} (AIR) dataset was manually labelled and consists of over 7K videos, 10 adverbs and 48 actions. 

To summarise our contributions: i) we propose a more effective approach to learn adverbs in videos. Our method achieves state-of-the-art results on multiple datasets and with fewer assumptions; ii) we introduce the AIR dataset for adverb recognition. 
Our focus on a domain where action changes are prominent and our careful manual annotation makes AIR more suitable for training and evaluating models for adverb understanding. We publicly release AIR and our code at \href{https://github.com/dmoltisanti/air-cvpr23}{github.com/dmoltisanti/air-cvpr23}.

\begin{figure*}
    \centering
    \includegraphics[width=\textwidth]{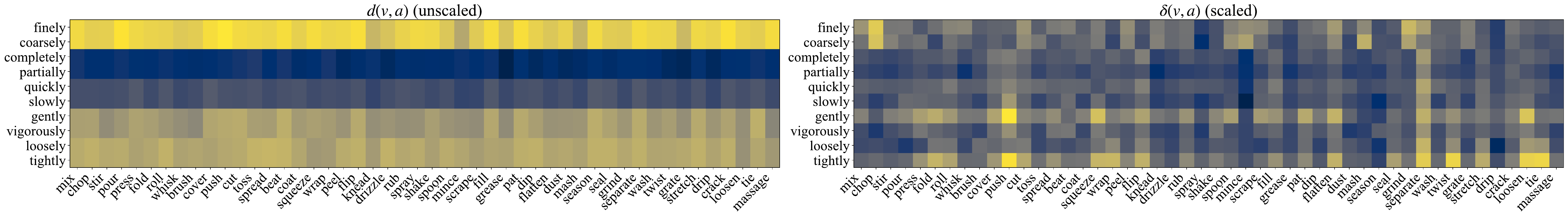}
    \caption{Comparison between the unscaled distance $d$ (left) and the scaled distance $\delta$ (right), which is scaled using the verb-adverb cosine similarity. Plotted values are the Euclidean distances between corresponding positive and negative sentences. Blue/yellow indicate small/large distances, i.e. a small/big change when adverbs are flipped in the positive sentence (see text for more information). 
    }
    \label{fig:dva}
    \vspace{-5pt}
\end{figure*}

\vspace{-7pt}
\section{Related Work}
\label{sec:related_work}

\fparagraph{Adverb Recognition} 
Doughty \textit{et al.}~\cite{doughty2020action} introduced the adverb recognition task and proposed Action Modifiers. Adverbs are treated as learnable parameters that guide the optimisation of the video embeddings via a triplet loss. 
We show that such approach can be difficult to optimise and propose an alternative learning method. 
Rather than modelling action changes as trainable parameters we define them measuring distances in a text embedding space. We then learn action modifications 
via regression. We provide a more technical comparison in Section~\ref{sec:method}.
A follow-up work~\cite{doughty2022you} 
proposes a pseudo-labelling method where videos are assigned new labels based on the model prediction. 
Pseudo labels are obtained in an adaptive away, i.e. the threshold determining whether a label should be assigned 
is adapted for each adverb independently. In~\cite{doughty2022you} the underlying model is also Action Modifiers~\cite{doughty2020action}. 
Pang \textit{et al.}~\cite{pang2018human, pang2020further} introduced the ``Behaviour Adverb'' task. Here actions are not goal-oriented (e.g. ``kiss, smoke'') and adverbs correspond to moods (e.g. ``sadly, solemnly''). 
Models use additional modalities such as human pose~\cite{pang2018human} or explicitly learn human expressions~\cite{pang2020further, fan2016video}. While related, this task is different from the scope of this work, where we aim to learn action changes in goal-oriented (e.g. cooking) videos. 

\vspace{-12pt}
\fparagraph{Adverb Datasets} 
HowTo100M Adverbs~\cite{doughty2020action} 
collects data from the instructional dataset HowTo100M~\cite{miech19howto100m} finding 
adverbs in the narration captions. Videos 
are loosely trimmed 
around the timestamp associated with the adverb. Due to the video-text misalignment in HowTo100M, and because training videos were not manually reviewed, 
videos in HowTo100M Adverbs are noisy. 
Authors estimated that only 44\% of the training videos actually show the action. 
In total 6 adverbs were annotated. 
Three other datasets were introduced in~\cite{doughty2022you}: ActivityNet/MSR-VTT/VATEX Adverbs. These are subsets of the namesake captioning datasets~\cite{krishna2017dense,xu2016msr,wang2019vatex}. 
Videos 
were also
obtained finding adverbs in captions, 
but annotations here are clean because captions were provided by annotators. 
We argue that the captioning nature of the original datasets is a reason of concern for learning action modifications. 
Indeed, 
adverbs tend to be a complementary description of the video, 
i.e. actions do not appear particularly influenced by the adverb. Peering into these datasets (see Figure~\ref{fig:datasets}), we found for example ``sit inside/outside, talk outdoors/indoors, 
walk in/out, move down/up'', etc. These adverbs modify the appearance of the video, however the action themselves are not modified. Arguably, the act of sitting is the same whether it takes place inside or outside, as is the act of 
walking in or out.
Lastly,~\cite{pang2018human, pang2020further} annotated adverbs for non-task-oriented actions, e.g. ``run, kiss'', however adverbs here express 
moods and manners, 
e.g. ``politely, 
reluctantly''. 

The limited availability of datasets specifically designed for adverb understanding motivates us to collect a new dataset better suited for this task. We focus on recipe instructional videos where actions change significantly according to the adverb. We collect Adverbs in Recipes, which contains 7K videos, 10 adverbs and 48 actions. We employ annotators to verify that actions are carried out as indicated by the adverb. We present our dataset in Section~\ref{sec:our_dataset}.

\vspace{-12pt}
\fparagraph{Video-text Retrieval} Understanding adverbs in videos is related to video-text retrieval~\cite{anne2017localizing,krishna2017dense,gao2017tall,mithun2018learning,liu2019use,wang2019language,dong2019dual,miech19endtoend,mun2020local,zeng2020dense,yang2021taco}. 
Theoretically, these methods could be used to retrieve videos using adverbs 
as text queries.
However, as noted in~\cite{doughty2022you}, most approaches encode text embeddings for whole sentences, 
i.e. it is hard to retrieve videos querying single words. Some works propose fine-grained video-text retrieval where single parts-of-speech can be queried~\cite{xu2015jointly,yu2017end,wray2019fine,chen2020fine}, however these works focus on verbs and nouns and were not evaluated for adverbs. As discussed before, adverbs are more abstract 
compared to verbs and nouns and are difficult to spatio-temporally localise. 
This entails that 
a video-text retrieval approach is unlikely to work well to find adverbs in videos. 
We test a retrieval baseline with an S3D model~\cite{miech19endtoend} where video and text embeddings were jointly learnt. Despite the strong performance 
on a number of tasks, we show that indeed this model struggles to achieve satisfactory performance. 

\begin{figure*}[t]
    \centering
    \includegraphics[width=0.78\textwidth]{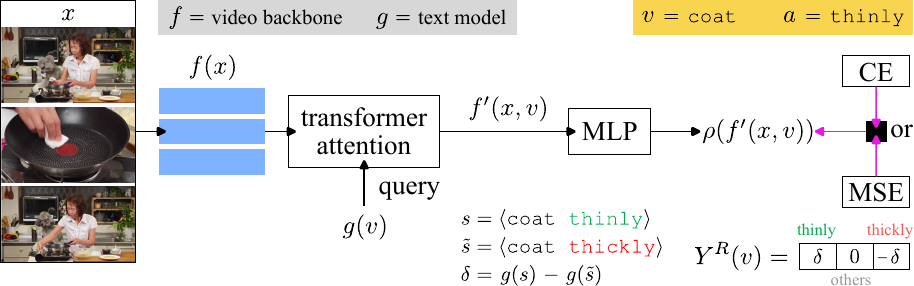}
    \caption{Pipeline of our method. Given a long video $x$ labelled with a verb $v$ and an adverb $a$ we learn video embeddings $f'(x, v)$ through attention. We optimise the model with two alternative methods: standard classification (CE: Cross Entropy) or regression (MSE: Mean Squared Error). We build a regression target measuring distances in a text embedding space, which estimates the action change we aim to learn in the video. The video backbone and text model are initialised from pre-trained models and are not fine-tuned during learning.}
    \label{fig:method}
    \vspace{-10pt}
\end{figure*}

\vspace{-3pt}
\section{Method}
\label{sec:method}
\vspace{-2pt}

We are given a video $x$ labelled with a verb $v$, which annotates the action, and an adverb $a$, which indicates the way the action is carried out. Our goal is to predict $a$ in $x$ given $v$. We also have 
a mapping $h$ between adverbs opposite in meaning, e.g. $h(\texttt{quickly}) = \texttt{slowly}$.  We use a pre-trained video backbone $f$ to extract video features $f(x)$. In this work we deal with long videos (10--40s), so we employ transformer-style attention~\cite{vaswani2017attention} to attend to the relevant parts of the video. To this end, we use a pre-trained text model $g$ to extract a text embedding $g(v)$, which we use as query for the attention, obtaining the projected video embedding $f'(x, v)$. To learn more specialised video features we feed $f'(x, v)$ to a shallow MLP and 
obtain the final output of the model $\rho(f'(x, v))$.
We propose two methods to optimise the network casting the problem either as a classification or a regression task. In the former case we utilise the standard Cross Entropy (CE) loss using the ground truth adverb label $a$. 
The CE 
pushes up predictions for the 
positive class and treats all negative classes equally. In some cases, however, it may be desirable 
to penalise the model more aggressively for incorrect predictions of antonym classes. 
Action Modifiers~\cite{doughty2020action} uses the triplet loss for this purpose. However, in~\cite{doughty2020action} negative samples are formed by only pairing antonyms. This is effective for distinguishing opposite adverbs, but does not train the model to distinguish an adverb from other negative non-antonym classes.

\vspace{15pt}

We thus propose a second approach based on regression. 
Casting the task as a regression problem we aim to directly learn the change the adverb introduces to the action and the video. However, 
we do not have a ground truth quantifying such action modifications. 
We use a text embedding space to estimate action 
changes. The high level idea is that we first build a minimal sentence $s$ representing the modified action. We similarly build a negative sentence $\tilde{s}$ representing the action modified in the opposite way.
Calculating the difference between $s$ and $\tilde{s}$ we have a proxy measure for the change the adverb applies to the action. We build $s$ concatenating the verb and the adverb (e.g. \texttt{coat \textcolor{Green}{thinly}}), then generate $\tilde{s}$ replacing the 
adverb with its antonym (e.g. \texttt{coat \textcolor{red}{thickly}}). 
We then extract text embeddings $g(s)$ and $g(\tilde{s})$. 
Note that $g$ receives a variable length sentence and outputs a fixed-dimension embedding. 
Let $\langle \rangle$ denote concatenation, so $s = \langle v, a \rangle$ and $\tilde{s} = \langle v, h(a) \rangle$. We 
measure the difference between $s$ and $\tilde{s}$ to capture how $a$ changes $v$:

\begin{equation}
    d(v, a) = \Big\Vert g\big(\langle v, a \rangle\big) - g\big(\langle v, h(a)\rangle\big) \Big\Vert _2
    \label{eq:distance}
\end{equation} 

Ideally, $d$ could be used as is. 
However, we observed that for a given adverb $a$ we obtain similar $d(v, a)$ across most verbs $v$. This defeats the purpose of measuring action changes via $d$ since $d$ itself is not discriminative enough. This 
happens 
because $d$ ignores the correlation between the verb and the adverb. Intuitively, if a verb and an adverb are not semantically correlated (e.g. ``run \textit{thinly}'') then we cannot expect to capture a meaningful change 
when flipping the adverb in the negative sentence  (recall that $g$ receives full sentences).
To address this issue we scale $d$ with the cosine similarity between the verb $v$ and the adverb $a$:

\vspace{-15pt}
\begin{align}
    \delta(v, a) = d(v, a) \cdot \dfrac{g(v) \cdot g(a)}{\Vert g(v) \Vert _2 \cdot \Vert g(a) \Vert _2}
\end{align}

\noindent Figure~\ref{fig:dva} compares $d$ (left) to $\delta$ (right) for a few (verb, adverb) combinations. 
Note how the the unscaled distance $d$ is not discriminative given that adverbs display a similar $d$ across verbs. On the other hand, the scaled $\delta$ shows a meaningful distance. For example, we observe a large $\delta$ for adverbs ``finely'' and ``coarsely'' when paired with verbs ``chop, mince, mash, grate''. 

Now that we have an estimate to measure action changes we can build a regression target:
\begin{equation}
Y^{R}(v) = (t_i,\forall i \in \{1, 2, \dots, A\}) =  
    \begin{cases}
       \phantom{-}\delta(v, a)& \hspace{-7pt}\text{\small if} \ y_i = a \\
       -\delta(v, a)& \hspace{-7pt} \text{\small if} \ y_i = h(a) \\
       \phantom{-}0& \hspace{-7pt}\text{\small otherwise}  
    \end{cases}
    \label{eq:reg_target}
\end{equation}

\noindent where $A$ is the number of adverbs in a dataset and $y_i$ denotes the $i$-th adverb. 
Let $\hat{y}_i^v$ indicate the output of the model for the $i$-th adverb obtained for verb $v$, i.e. querying the attention model with $g(v)$.
The full output of the model is then $\hat{Y}^v = \rho (f'(x, v)) = (\hat{y}_i^v \ , \forall i \in \{1, 2, \dots, A\})$. 

We use the Mean Squared Error to train the network: $\mathcal{L}^{R} = \Vert Y^{R}(v) - \hat{Y}^v \Vert _2$.
The loss pushes opposite adverbs furthest apart (due to the negative sign of $\delta$ for the antonym). 
The loss also tells the model that the video is not modified by other unrelated adverbs (target is 0 for classes that are neither ground truth nor antonyms). This discourages the model to predict negative non-antonym adverbs.
Figure~\ref{fig:method} illustrates our method. Note that the video backbone and text model are not part of the training, i.e. they are 
initialised from pre-trained models and are not fine-tuned.

\vspace{-10pt}

\fparagraph{Learning without Antonyms} We also consider a setting where adverbs are not paired in antonyms. Instead of replacing the ground truth 
adverb with its antonym we simply remove the adverb and generate $\tilde{s} = v$. In this case $\delta$ measures the difference between the action modified by the adverb and the unmodified version of the action. $Y^R$ is then $\delta$ for the ground truth adverb and 0 for all other adverbs. 

\vspace{-10pt}

\fparagraph{Inference} As in~\cite{doughty2020action} we are given the action label $v$ during testing. We query the attention model with $g(v)$ and use $\hat{Y}^v$ 
to predict 
each adverb.
We also test the model without action labels. 
In such case we obtain $f'(x, v)$
querying all actions 
and take the maximum prediction for each adverb: 

\vspace{-10pt}
\begin{equation}
p(x) = (\max\limits_{v}(\hat{y}_i^v \ , \ \forall v \in \mathcal{V})\ , \ \forall i \in \{1,2,\dots,A\})
\end{equation}

\noindent where $\mathcal{V}$ denotes the set of action labels. 

\paragraph{Comparison to Action Modifiers} Our framework shares the same design  as Action Modifiers~\cite{doughty2020action} until $f'(x, v)$. In~\cite{doughty2020action} $f'(x, v)$ is optimised with two triplet losses: an action loss to help the attention model and an adverb loss to learn adverbs. 
We do not use an additional action loss. 
In Action Modifiers adverbs are also modelled as action changes utilising text embeddings. However, in~\cite{doughty2020action} such change is treated as an additional parameter to be learnt. Specifically, an adverb is represented as a learnable matrix $W_a$. The linear combination $o_v^a = W_a g(v)$ represents the change the adverb $a$ applies to the verb $v$, where $g(v)$ is a text embedding as in our model. Importantly, $o_v^a$ is used in the triplet losses, thus the optimisation of the video embedding $f'(x, v)$ depends on the optimisation of parameters $W_a$. We argue and show that learning adverbs with this approach  
can be difficult as the action change the model looks for in the video has also to be learnt. In our formulation instead the 
action change is pre-computed and is not learnt. 
We also provide explicit penalty for non-antonym classes rather than contrasting only opposite adverbs.
Comparing capacity, our model has a smaller number of parameters which scales better with the number of adverbs. Indeed, each adverb in Action Modifiers requires a learnable matrix, whereas in our model only the last layer of the MLP varies with the number of adverbs (see Appendix~\ref{sec:capacity}). 

\vspace{-7pt}
\section{Adverbs in Recipes: the AIR Dataset}
\label{sec:our_dataset}

\begin{table}[t]
\centering
\resizebox{\columnwidth}{!}{%
\begin{tabular}{@{}llclcccc@{}}
\toprule
                    & Dataset                   & Accuracy  & Duration  & Adv & Act & Pairs & Videos \\ \midrule
\multirow{3}{*}{\rotatebox{90}{\textcolor{captioning_col}{CAPT}}}  & ActivityNet Adverbs~\cite{doughty2022you}       & 89.0\%   & 37.5s         & 20      & 114     & 635   & 3,099   \\
                        & MSR-VTT Adverbs~\cite{doughty2022you}           & 91.0\%  & 15.7s         & 18      & 106     & 450   & 1,824   \\
                        & VATEX Adverbs~\cite{doughty2022you}             & 93.5\% & 10.0s (f) & 34      & 135     & 1,524 & 14,617   \\ \midrule
\multirow{2}{*}{\rotatebox{90}{\textcolor{instr_col}{INST}}} & HowTo100M Adverbs~\cite{doughty2020action}         & 44.0\%  & 20.0s (f) & 6       & 72      & 257   & 5,824   \\
                        & \textbf{Adverbs in Recipes} (ours) & 95.3\%   & 8.4s          & 10      & 48      & 186   & 7,003   \\ \bottomrule 
\end{tabular}%
}
\caption{Comparing video datasets for adverb recognition. \textcolor{captioning_col}{CAPT}: captioning datasets, where adverbs are 
descriptive. \textcolor{instr_col}{INST}: instructional datasets, where adverbs are action-focused. ``Accuracy'' indicates whether the action is visible as indicated by the adverb. ``Duration'' reports the average length, where (f) denotes a fixed duration.
``Pairs'' counts  
(verb, adverb) appearing in the dataset rather then the Cartesian product of all verbs and adverbs.} \vspace{-15pt}
\label{tab:datasets}
\end{table}

In Section~\ref{sec:related_work} we motivated the need for a new dataset for adverb recognition. Here we introduce the Adverbs in Recipes (AIR) dataset. 
We wish to find a set of videos where actions change significantly according to the adverb. Compared to captioning benchmarks, instructional datasets contain videos where we can expect to find more of such actions.
For this reason we source AIR from HowTo100M~\cite{miech19howto100m}. 
We restrict our interest to recipe videos since changes in cooking actions 
play an important role. For example, a carrot sliced \textit{\`a la julienne} (thinly) looks very different from a carrot chopped in thick dices (coarsely). 

In other words, adverbs are ``easily visible'' in recipe videos and involve interesting visual cues 
such as speed, temporal/spatial completeness and objects end-states.

\fparagraph{Data Collection} We start selecting all recipe videos in HowTo100M. 
We then parse captions to keep only those containing a verb and an adverb. 
We filter videos removing verbs that indicate non visual actions (e.g. ``watch, wait'') or very long-term actions (e.g. ``rise, grow''), since we do not expect these changes to be fully visible in the videos. We manually merge similar verbs, obtaining in total 48 verbs.   
We also remove infrequent and too generic adverbs (e.g. ``carefully''). We cluster similar adverbs gathering a total of 10 adverbs, which we pair in antonyms. 
To trim videos in a tight manner 
we use 
the timestamps of the first/last word in a caption as the start/end of the action segment. At this point 
we have a set of videos of varying duration, each accompanied with a verb and an adverb. 
We ask annotators on Amazon Mechanical Turk (AMT) to check whether the action in a video is visible and is carried out as indicated by the adverb. For robustness we ask 3 different annotators to check the same video and employ 5 people for edge cases where annotators disagreed. 
We keep videos where the majority of the annotators confirmed that the action is visible as indicated by the adverb, collecting 7,003 videos. 
Figure~\ref{fig:dva} shows the verbs (columns) and adverbs (rows) in AIR.
We provide more details in Appendix~\ref{sec:air_details}. 

\fparagraph{Comparison to Other Datasets} Table~\ref{tab:datasets} compares AIR to other adverb datasets\footnote{
We report the number of videos from~\cite{doughty2022you}. Some videos are missing as they were removed from YouTube.
We were able to download: ActivityNet A.: 2,972 - MSR-VTT A.: 1,747 - VATEX A.: 13,947, HowTo100M A.: 5,462. Duration and pairs are reported from the downloaded videos.} reviewed in Section~\ref{sec:related_work}. 
``Accuracy'' estimates the percentage of videos where the action and the adverb effect are visible. 
On the existing datasets this was estimated watching 200 videos in~\cite{doughty2022you}. For AIR multiple annotators reviewed all videos, 
however, annotations on AMT can be noisy. We review 302 videos (5\% of the dataset) and report this check as ``Accuracy''. Compared to the instructional-based HowTo100M Adverbs, AIR contains over 1,100 additional clips and 4 more adverbs. Importantly, our videos are manually reviewed (AIR accuracy is over 95\% compared to 44\%) and better trimmed (averaged duration is 8s compared to fixed 20s segments).
AIR features 48 actions which is the smallest number among the five datasets (and consequently the smallest number of pairs). This is due to our more restrictive verb filtering based on visual and task-oriented actions. For example, in the captioning datasets there are actions such as ``tell, talk, sing, look''. 
Actions in HowTo100M Adverbs are task oriented, however there are also verbs entailing long-term actions (e.g. ``brew''), which are unlikely to be fully visible. 

\begin{figure}[t]
    \centering
    \includegraphics[width=1\columnwidth]{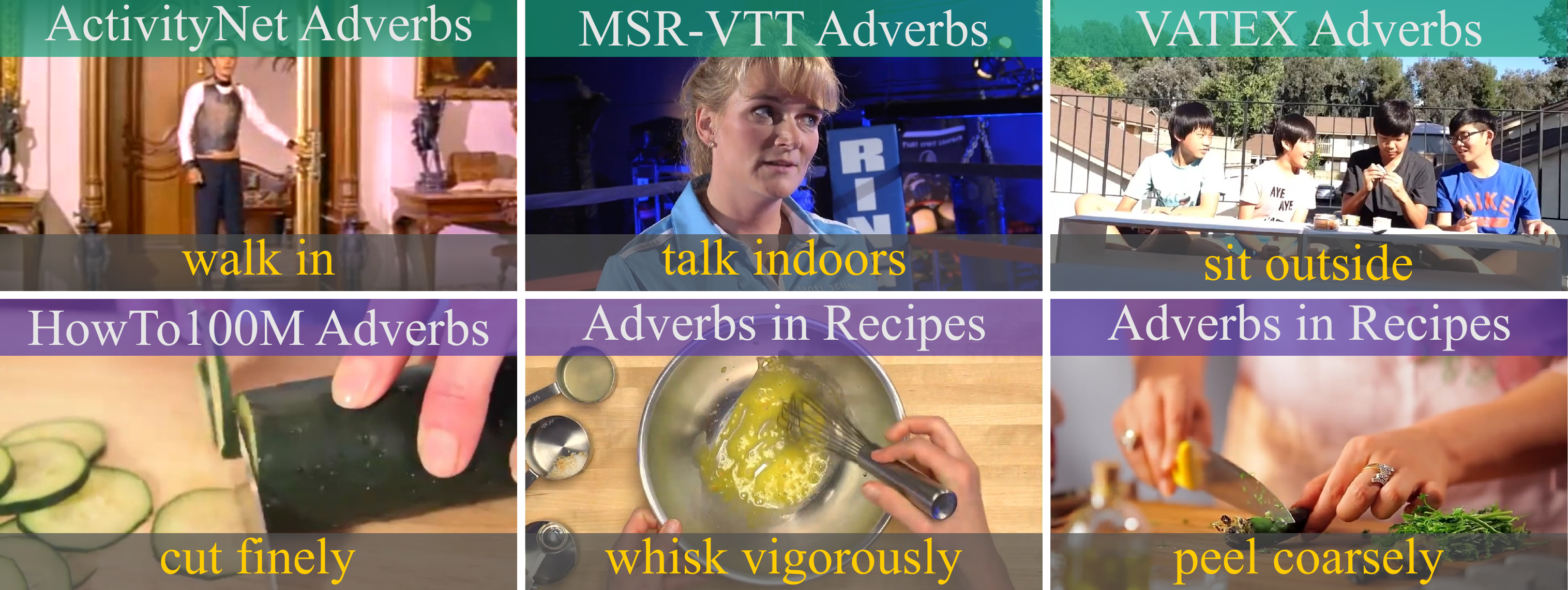}
    \caption{Samples from existing datasets for adverb recognition and our new AIR dataset. Many adverbs in the \textcolor{captioning_col}{captioning datasets} appear to be  
    descriptive rather than a key modification of the action (e.g. the act of talking/sitting is the same whether it takes place indoors/outdoors). In contrast, adverbs in the \textcolor{instr_col}{instructional datasets} apply a more prominent change in the action and its outcome (e.g. a vegetable peeled coarsely will still have some skin attached).}
    \label{fig:datasets}
    \vspace{-5pt}
\end{figure}

Figure~\ref{fig:datasets} shows a few samples from AIR as well as the other datasets. Note how adverbs in the instructional datasets apply a key modification in the action, whereas for the captioning datasets adverbs are 
descriptive. 
Captions in the 
the captioning datasets 
are reliable because they were provided by human annotators. However, annotators did not necessarily use adverbs to indicate a modification to the action. Thus, due to the automatic processing of these datasets, there are several cases where adverbs are descriptive and do not influence the action.
Naturally, Figure~\ref{fig:datasets} is not meant to be an exhaustive representation of the captioning datasets from~\cite{doughty2022you}. 
These datasets also contain action-focused adverbs, 
however without extensively reviewing the videos and the associated adverbs it is hard to gauge how many actions are effectively modified by the adverb.

To summarise the contributions of AIR: i) we employ annotators to verify that actions are performed in the way indicated by the adverb; ii) this allowed us to trim videos more tightly and obtain cleaner clips with less unrelated content; iii) we carefully choose verbs and adverbs to pick videos where action changes are significant. 
While we restrict our domain to recipes, cooking actions are visually diverse, 
entail complex object interactions and are particularly affected by the nuanced ways they are carried out. 

\section{Experiments}
\label{sec:experiments}

\begin{table*}[t]
\centering
\resizebox{\textwidth}{!}{%
\begin{tabular}{@{}lc@{\hspace{1cm}}ccc|cccc@{\hspace{1cm}}ccc|ccc|ccc@{}}
\cmidrule{3-8} \cmidrule{10-18}
 & &
  \multicolumn{6}{c}{Instructional datasets. Action-focused adverbs} & &
  \multicolumn{9}{c}{Captioning datasets. Descriptive adverbs} \\ \cmidrule{3-8} \cmidrule{10-18}
 & &
  \multicolumn{3}{c|}{HowTo100M Adverbs~\cite{doughty2020action}} &
  \multicolumn{3}{c}{Adverbs in Recipes} & &
  \multicolumn{3}{c|}{ActivityNet Adverbs~\cite{doughty2022you}} &
  \multicolumn{3}{c|}{MSR-VTT Adverbs~\cite{doughty2022you}} &
  \multicolumn{3}{c}{VATEX Adverbs~\cite{doughty2022you}} \\ \cmidrule{3-8} \cmidrule{10-18}
  & &
  mAP W &
  mAP M &
  Acc-A &
  mAP W &
  mAP M &
  Acc-A & &
  mAP W &
  mAP M &
  Acc-A &
  mAP W &
  mAP M &
  Acc-A &
  mAP W &
  mAP M &
  Acc-A \\ \cmidrule{1-1} \cmidrule{3-8} \cmidrule{10-18}
Priors & &
  0.446 &
  0.354 &
  0.786 &
  0.491 &
  0.263 &
  0.854 & &
  \textbf{0.217} &
  \textbf{0.159} &
  0.745 &
  \textbf{0.308} &
  \textbf{0.152} &
  0.723 &
  0.216 &
  0.086 &
  0.752 \\
S3D pre-trained & &
  0.339 &
  0.238 &
  0.560 &
  0.389 &
  0.173 &
  0.735 & &
  0.118 &
  0.070 &
  0.560 &
  0.194 &
  0.075 &
  0.603 &
  0.122 &
  0.038 &
  0.586 \\
Act Mod~\cite{doughty2020action} & &
  0.406 &
  0.372 &
  0.796 &
  0.509 &
  0.251 &
  0.857 & &
  0.184 &
  0.125 &
  \textbf{0.753} &
  0.233 &
  0.127 &
  0.731 &
  0.139 &
  0.059 &
  0.751 \\
MLP + Act Mod~\cite{doughty2020action} & &
  0.279 &
  0.193 &
  0.793 &
  0.382 &
  0.144 &
  \textbf{0.860} & &
  0.131 &
  0.087 &
  \textbf{0.753} &
  0.184 &
  0.123 &
  0.731 &
  0.209 &
  0.096 &
  0.752 \\
\cellcolor[HTML]{B7FBFF}CLS & &
  \cellcolor[HTML]{B7FBFF}\textbf{0.562} &
  \cellcolor[HTML]{B7FBFF}0.420 &
  \cellcolor[HTML]{B7FBFF}0.786 &
  \cellcolor[HTML]{B7FBFF}0.606 &
  \cellcolor[HTML]{B7FBFF}\textbf{0.289} &
  \cellcolor[HTML]{B7FBFF}0.841 & &
  \cellcolor[HTML]{B7FBFF}0.130 &
  \cellcolor[HTML]{B7FBFF}0.096 &
  \cellcolor[HTML]{B7FBFF}0.741 &
  \cellcolor[HTML]{B7FBFF}0.305 &
  \cellcolor[HTML]{B7FBFF}0.131 &
  \cellcolor[HTML]{B7FBFF}0.751 &
  \cellcolor[HTML]{B7FBFF}\textbf{0.283} &
  \cellcolor[HTML]{B7FBFF}\textbf{0.108} &
  \cellcolor[HTML]{B7FBFF}0.754 \\
\cellcolor[HTML]{FFE0D1}REG - fixed $\delta$ & &
  \cellcolor[HTML]{FFE0D1}0.320 &
  \cellcolor[HTML]{FFE0D1}0.215 &
  \cellcolor[HTML]{FFE0D1}0.706 &
  \cellcolor[HTML]{FFE0D1}0.554 &
  \cellcolor[HTML]{FFE0D1}0.193 &
  \cellcolor[HTML]{FFE0D1}0.837 & &
  \cellcolor[HTML]{FFE0D1}0.115 &
  \cellcolor[HTML]{FFE0D1}0.075 &
  \cellcolor[HTML]{FFE0D1}0.706 &
  \cellcolor[HTML]{FFE0D1}0.203 &
  \cellcolor[HTML]{FFE0D1}0.100 &
  \cellcolor[HTML]{FFE0D1}0.706 &
  \cellcolor[HTML]{FFE0D1}0.175 &
  \cellcolor[HTML]{FFE0D1}0.051 &
  \cellcolor[HTML]{FFE0D1}0.701 \\
\cellcolor[HTML]{FFECA8}REG & &
  \cellcolor[HTML]{FFECA8}0.555 &
  \cellcolor[HTML]{FFECA8}\textbf{0.423} &
  \cellcolor[HTML]{FFECA8}\textbf{0.799} &
  \cellcolor[HTML]{FFECA8}\textbf{0.613} &
  \cellcolor[HTML]{FFECA8}0.244 &
  \cellcolor[HTML]{FFECA8}0.847 & &
  \cellcolor[HTML]{FFECA8}0.119 &
  \cellcolor[HTML]{FFECA8}0.079 &
  \cellcolor[HTML]{FFECA8}0.714 &
  \cellcolor[HTML]{FFECA8}0.282 &
  \cellcolor[HTML]{FFECA8}0.114 &
  \cellcolor[HTML]{FFECA8}\textbf{0.774} &
  \cellcolor[HTML]{FFECA8}0.261 &
  \cellcolor[HTML]{FFECA8}0.086 &
  \cellcolor[HTML]{FFECA8}\textbf{0.755} \\ \cmidrule{1-1} \cmidrule{3-8} \cmidrule{10-18}
\end{tabular}%
}
\caption{Results obtained using the action label during inference.  mAP W/M: mean Average Precision with weighted (W) and macro (M) averaging. Acc-A: adverb-vs-antonym accuracy. Coloured rows indicate variants of our method. Bold denotes best result per column.  
In instructional datasets (left) adverbs are action-focused, so these are more reliable benchmarks to learn action changes. In captioning datasets (right) adverbs are more descriptive and do not influence the action significantly. 
As such, these datasets are less reliable. 
\vspace{-10pt}
}
\label{tab:results_all}
\end{table*}

\fparagraph{Implementation Details}
We use an S3D model~\cite{xie2018rethinking,miech19endtoend} jointly pre-trained on video and text on HowTo100M~\cite{miech19howto100m} as our video backbone\footnote{There is no official train/test split in HowTo100M, thus S3D in~\cite{miech19endtoend} was trained on videos that appear in the test sets of HowTo100M Adverbs and AIR. However, this is not a concern: Tables~\ref{tab:results_all},  \ref{tab:results_no_act_gt} show that the S3D pre-trained baseline achieves poor performance, as the original S3D training objective (video-text alignment) is substantially different from ours. Importantly, all methods
receive the same features for fair comparison.}. Specifically, we use the video output \texttt{mixed\_5c} to obtain $f(x)$. Following~\cite{doughty2020action} we split videos in 1-second segments and sample 16 RGB frames from each segment to obtain the video features. 
Stacking features from the video segments we have $f(x)^{T \times D}$, where $T$ varies depending on the dataset and $D$ is 1024. We use 4 heads for the attention model. 
The dimension $E$ of the query, key and value is 512, thus the projected video embedding $f'(x, v)$ has dimension $1 \times E$. 
The MLP in our model has 3 layers (hidden units of dimension 512) with ReLU activation function for the hidden layers and dropout~\cite{srivastava2014dropout} set to 0.1. 
We train all models for 1000 epochs with the ADAM optimiser~\cite{kingma2015adam}, setting learning rate to $1e^{-4}$, weight decay to $5e^{-5}$ and batch size to 512. We use the text model jointly trained with S3D to extract text embeddings $g(.)$. This network, like the visual backbone, is not fine-tuned. We refer to~\cite{miech19endtoend} for more details about S3D and the text model. $f(x)$ and $g(.)$ are the same for all experiments.

\vspace{-12pt}

\fparagraph{Baselines} \textit{Priors:} here we do not train any model and simply use the prior distributions of the training set to make a prediction. When using the action label during testing, the prediction for an adverb is the number of training samples labelled with the adverb and the given verb, i.e. $p(x, v) = (\pi(v, a) \ , \ \forall a \in \mathcal{A})$, where $\pi$ denotes the frequency of $(v, a)$ in the training set and $\mathcal{A}$ is the set of adverbs in a dataset. While this baseline is a simple look-up table, it can achieve strong performance since it exploits the co-occurrence of verbs and adverbs. When the action label is not used during testing the priors baseline predicts the frequency of each adverb in the training set, i.e. $p(x) = (\pi(a) \ , \ \forall a \in \mathcal{A})$.

\textit{S3D pre-trained:} we test the S3D model we use as the video-text backbone 
for all experiments. 
Video and text embeddings were jointly learnt in this model, so we use the dot product between $f(x)$ and $g(.)$ for predictions. When the action label $v$ is given during inference the prediction is $p(x, v) = (f(x) \cdot g( \langle v, a \rangle) \ , \  \forall a \in \mathcal{A})$, where $g(\langle v, a\rangle)$ is the text embedding obtained concatenating the given $v$ and each adverb $a$. When the action label is not given, we have $p(x) = (\max\limits_{v}(f(x) \cdot g(\langle v, a\rangle) \ , \  \forall v \in \mathcal{V}) \ , \ \forall a \in \mathcal{A})$, where $\mathcal{V}$ is the set of verbs. This S3D instance has shown very strong performance on several tasks~\cite{miech19endtoend}, thus this baseline helps gauging 
the difficulty of detecting adverbs following a basic video-text retrieval approach.

\vspace{-12pt}

\fparagraph{SOTA}\textit{Act Mod:} we compare against the state-of-the-art for adverb recognition, Action Modifiers~\cite{doughty2020action}, 
both using the original model and a deeper version with an additional MLP (\textit{MLP + Act Mod}). Here we append an identical MLP as in our model to the output of the transformer attention. This variant allows a better comparison with our method. 
We use the official code implementation to run our experiments. Note that Action Modifiers results reported here are not directly comparable to those reported in~\cite{doughty2020action, doughty2022you}. This is because we could not download several videos since they were removed from YouTube. Also, we use 
S3D features (identical for all experiments) instead of I3D~\cite{carreira2017quo} features. 

\vspace{-13pt}

\fparagraph{Our Variants}\textit{CLS:} our model trained with classification. \textit{REG - fixed $\delta$}: our model trained with regression discarding text embeddings altogether, setting $\delta=1$ in Equation~\ref{eq:reg_target}. This is to validate the premise of using textual context to build a regression target.
\textit{REG}: our model trained with regression using the full formulation.

\vspace{-13pt}

\fparagraph{Evaluation Metrics}
We report mean Average Precision using two types of averaging: i) weighted (mAP W), where class scores are weighted according to the support size (smaller classes have a smaller weight); ii) macro (mAP M), where all classes have equal weight, which corresponds to ``adverb-to-video (all)'' in~\cite{doughty2020action}. All the evaluated datasets exhibit a significant class imbalance, thus mAP M is a stricter metric compared to mAP W. 
We also report binary antonym accuracy (Acc-A), which corresponds to ``video-to-adverb (antonym)'' in~\cite{doughty2020action}. Here we only look at the prediction of an adverb $a$ versus its antonym $h(a)$. A prediction $p$ is correct if $p(a) > p(h(a))$. All metrics are computed for adverb classes.
To show the full potential of all methods we report each best metric independently, i.e. results may come from different epochs. 
This is to provide a robust comparison, however we note that all models reach stable convergence. 

\vspace{-13pt}

\fparagraph{Datasets}
We present experiments on the two instructional datasets we saw earlier: HowTo100M Adverbs~\cite{doughty2020action} and our Adverbs In Recipes. As discussed before, we argue that instructional videos are more suitable for this task as adverbs are action-focused, i.e. actions change significantly according to the adverb. For completeness, we also evaluate the three adverb datasets sourced from captioning videos in~\cite{doughty2022you}: ActivityNet/MSR-VTT/VATEX Adverbs\footnote{For these datasets there are several partitions for different tasks (e.g. unseen domain). For convenience we create our own train/test split.}. Adverbs here are  
descriptive, thus 
these datasets are less reliable for evaluation.
Datasets are summarised in Table~\ref{tab:datasets}. 

\vspace{-3pt}
\subsection{Results}

\fparagraph{Overview} We begin our discussion looking at results in Table~\ref{tab:results_all} on the instructional datasets (left).
Here we achieve new state-of-the-art results on all metrics,
with the exception of Acc-A on AIR, where the contrastive loss in \textit{Act Mod}~\cite{doughty2020action} is slightly more effective when considering only antonyms. Indeed, the gap between our best variant 
and \textit{Act Mod} is more noticeable for the more challenging mAP W/M metrics. 
In~\cite{doughty2020action} negative samples are formed only pairing opposite adverbs, thus the model does not receive enough penalty for other negative adverbs. In contrast, we 
provide 
explicit penalty for non-antonym negative adverbs. The poor performance of 
\textit{S3D pre-trained} 
indicates that learning adverbs in videos is not trivial. 
Adverbs are not well defined visual entities like objects or actions, thus video-text retrieval approaches looking for correlations between text and video embeddings are likely sub-optimal, even when such embeddings are jointly learnt on the same dataset with a strong model, as is the case for S3D. 

Still looking at the instructional datasets, on AIR all methods achieve higher mAP W/Acc-A but lower mAP M. Considering that there are more adverbs in AIR 
than in HowTo100M Adverbs, 
the higher mAP W/Acc-A is indicative of the quality of AIR: given that videos are tighter and cleaner it is possible to learn adverbs more effectively. 
The lower mAP M is due to the long-tail class distribution and  
shows the opportunities AIR offers to improve methods. 

On the captioning datasets we observe that the \textit{Priors} baseline outperforms or achieves high performance, especially on the mAP metrics. 
This supports our argument that captioning datasets are not particularly suitable for our task. 
If a simple look-up table can surpass learning-based methods, then we conclude it is very difficult to learn adverbs as action changes, 
because actions are not particularly influenced by adverbs here. While these datasets exhibit a skewed verb-adverb distribution, so do the instructional ones. In fact, on VATEX Adverbs (the biggest dataset with 1,524 pairs) our method attains marginally better results. This is to say that the problem does not lie in the imbalanced class distribution, rather in the 
nature of the videos. 

\begin{table*}[t]
\centering
\resizebox{\textwidth}{!}{%
\begin{tabular}{@{}lc@{\hspace{1cm}}ccc|cccc@{\hspace{1cm}}ccc|ccc|ccc@{}}
\cmidrule{3-8} \cmidrule{10-18}
 & &
  \multicolumn{6}{c}{Instructional datasets. Action-focused adverbs} & &
  \multicolumn{9}{c}{Captioning datasets. Descriptive adverbs} \\ \cmidrule{3-8} \cmidrule{10-18}
 & &
  \multicolumn{3}{c|}{HowTo100M Adverbs~\cite{doughty2020action}} &
  \multicolumn{3}{c}{Adverbs in Recipes} & &
  \multicolumn{3}{c|}{ActivityNet Adverbs~\cite{doughty2022you}} &
  \multicolumn{3}{c|}{MSR-VTT Adverbs~\cite{doughty2022you}} &
  \multicolumn{3}{c}{VATEX Adverbs~\cite{doughty2022you}} \\ \cmidrule{3-8} \cmidrule{10-18}
  & &
  mAP W &
  mAP M &
  Acc-A &
  mAP W &
  mAP M &
  Acc-A & &
  mAP W &
  mAP M &
  Acc-A &
  mAP W &
  mAP M &
  Acc-A &
  mAP W &
  mAP M &
  Acc-A \\ \cmidrule{1-1} \cmidrule{3-8} \cmidrule{10-18}
Priors & &
  0.247 &
  0.167 &
  0.560 &
  0.335 &
  0.100 &
  0.835 & &
  0.094 &
  0.050 &
  0.692 &
  0.137 &
  0.056 &
  0.723 &
  0.089 &
  0.029 &
  0.651 \\
S3D pre-trained & &
  0.329 &
  0.222 &
  0.594 &
  0.425 &
  0.177 &
  0.702 & &
  0.113 &
  0.065 &
  0.598 &
  0.199 &
  0.088 &
  0.603 &
  0.117 &
  0.037 &
  0.604 \\
Act Mod~\cite{doughty2020action} & &
  0.269 &
  0.183 &
  0.601 &
  0.356 &
  0.115 &
  0.835 & &
  0.110 &
  0.062 &
  0.716 &
  0.164 &
  0.071 &
  0.723 &
  0.100 &
  0.035 &
  0.686 \\
MLP + Act Mod~\cite{doughty2020action} & &
  0.268 &
  0.184 &
  0.706 &
  0.369 &
  0.115 &
  0.835 & &
  0.110 &
  0.062 &
  0.714 &
  0.163 &
  0.080 &
  0.723 &
  0.092 &
  0.036 &
  0.686 \\
\cellcolor[HTML]{B7FBFF}CLS & &
  \cellcolor[HTML]{B7FBFF}\textbf{0.404} &
  \cellcolor[HTML]{B7FBFF}\textbf{0.307} &
  \cellcolor[HTML]{B7FBFF}\textbf{0.724} &
  \cellcolor[HTML]{B7FBFF}\textbf{0.578} &
  \cellcolor[HTML]{B7FBFF}\textbf{0.314} &
  \cellcolor[HTML]{B7FBFF}0.841 & &
  \cellcolor[HTML]{B7FBFF}\textbf{0.129} &
  \cellcolor[HTML]{B7FBFF}\textbf{0.096} &
  \cellcolor[HTML]{B7FBFF}\textbf{0.741} &
  \cellcolor[HTML]{B7FBFF}\textbf{0.304} &
  \cellcolor[HTML]{B7FBFF}0.131 &
  \cellcolor[HTML]{B7FBFF}0.754 &
  \cellcolor[HTML]{B7FBFF}0.161 &
  \cellcolor[HTML]{B7FBFF}0.053 &
  \cellcolor[HTML]{B7FBFF}0.712 \\
\cellcolor[HTML]{FFE0D1}REG - fixed $\delta$ & &
  \cellcolor[HTML]{FFE0D1}0.320 &
  \cellcolor[HTML]{FFE0D1}0.215 &
  \cellcolor[HTML]{FFE0D1}0.706 &
  \cellcolor[HTML]{FFE0D1}0.554 &
  \cellcolor[HTML]{FFE0D1}0.193 &
  \cellcolor[HTML]{FFE0D1}0.837 & &
  \cellcolor[HTML]{FFE0D1}0.114 &
  \cellcolor[HTML]{FFE0D1}0.075 &
  \cellcolor[HTML]{FFE0D1}0.706 &
  \cellcolor[HTML]{FFE0D1}0.204 &
  \cellcolor[HTML]{FFE0D1}0.096 &
  \cellcolor[HTML]{FFE0D1}0.709 &
  \cellcolor[HTML]{FFE0D1}0.158 &
  \cellcolor[HTML]{FFE0D1}0.047 &
  \cellcolor[HTML]{FFE0D1}0.703 \\
\cellcolor[HTML]{FFECA8}REG & &
  \cellcolor[HTML]{FFECA8}0.377 &
  \cellcolor[HTML]{FFECA8}0.277 &
  \cellcolor[HTML]{FFECA8}\textbf{0.724} &
  \cellcolor[HTML]{FFECA8}0.520 &
  \cellcolor[HTML]{FFECA8}0.213 &
  \cellcolor[HTML]{FFECA8}\textbf{0.844} & &
  \cellcolor[HTML]{FFECA8}0.120 &
  \cellcolor[HTML]{FFECA8}0.079 &
  \cellcolor[HTML]{FFECA8}0.716 &
  \cellcolor[HTML]{FFECA8}0.276 &
  \cellcolor[HTML]{FFECA8}\textbf{0.133} &
  \cellcolor[HTML]{FFECA8}\textbf{0.774} &
  \cellcolor[HTML]{FFECA8}\textbf{0.169} &
  \cellcolor[HTML]{FFECA8}\textbf{0.057} &
  \cellcolor[HTML]{FFECA8}\textbf{0.737} \\ \cmidrule{1-1} \cmidrule{3-8} \cmidrule{10-18}
\end{tabular}%
}
\caption{Results obtained \textit{without} action labels during inference. mAP W/M: mean Average Precision with weighted (W) and macro (M) averaging. Acc-A: adverb-vs-antonym accuracy. Coloured rows indicate variants of our method. Bold denotes best result per column.}
\label{tab:results_no_act_gt}
\end{table*}

\begin{table*}[t]
\centering
\resizebox{0.9\textwidth}{!}{%
\begin{tabular}{@{}lcc@{\hspace{1cm}}cc|ccc@{\hspace{1cm}}cc|cc|cc@{}}
\cmidrule{4-7} \cmidrule{9-14}
 & & &
  \multicolumn{4}{c}{Instructional datasets. Action-focused adverbs} & &
  \multicolumn{6}{c}{Captioning datasets. Descriptive adverbs} \\ \cmidrule{4-7} \cmidrule{9-14}
 & & &
  \multicolumn{2}{c|}{HowTo100M Adverbs~\cite{doughty2020action}} &
  \multicolumn{2}{c}{Adverbs in Recipes} & &
  \multicolumn{2}{c|}{ActivityNet Adverbs~\cite{doughty2022you}} &
  \multicolumn{2}{c|}{MSR-VTT Adverbs~\cite{doughty2022you}} &
  \multicolumn{2}{c}{VATEX Adverbs~\cite{doughty2022you}} \\ \cmidrule{4-7} \cmidrule{9-14}
  &
  Antonyms & &
  mAP W &
  mAP M &
  mAP W &
  mAP M & &
  mAP W &
  mAP M &
  mAP W &
  mAP M &
  mAP W &
  mAP M \\ \cmidrule{1-2} \cmidrule{4-7} \cmidrule{9-14}
  Priors & \xmark & &
  0.446 &
  0.354 &
  0.491 &
  0.263 & &
  \textbf{0.217} &
  \textbf{0.159} &
  \textbf{0.308} &
  \textbf{0.152} &
  0.216 &
  0.086
  \\
S3D pre-trained & \xmark & &
  0.339 &
  0.238 &
  0.389 &
  0.173 & &
  0.118 &
  0.070 &
  0.194 &
  0.075 &
  0.122 &
  0.038 \\ \cmidrule{1-2}  \cmidrule{4-7} \cmidrule{9-14}
 & 
  \cmark & &
  0.406 &
  0.372 &
  0.509 &
  0.251 & &
  0.184 &
  0.125 &
  0.233 &
  0.127 &
  0.139 &
  0.059 \\
\multirow{-2}{*}{Act Mod~\cite{doughty2020action}} &
  \xmark & &
  0.408 &
  0.352 &
  0.508 &
  0.249 & &
  0.187 &
  0.127 &
  0.233 &
  0.134 &
  0.144 &
  0.060 \\ \cmidrule{1-2}  \cmidrule{4-7} \cmidrule{9-14} &
  \cmark & &
    0.279 &
  0.193 &
  0.382 &
  0.144 & &
  0.131 &
  0.087 &
  0.184 &
  0.123 &
  0.209 &
  0.096 
\\
\multirow{-2}{*}{MLP + Act Mod~\cite{doughty2020action}} &
  \xmark & &
    0.281 &
  0.198 &
  0.383 &
  0.140 & &
  0.136 &
  0.090 &
  0.193 &
  0.122 &
  0.213 &
  0.098 
   \\ \cmidrule{1-2}  \cmidrule{4-7} \cmidrule{9-14}
\cellcolor[HTML]{B7FBFF}CLS &
  \cellcolor[HTML]{B7FBFF}\xmark & &
  \cellcolor[HTML]{B7FBFF}0.562 &
  \cellcolor[HTML]{B7FBFF}0.420 &
  \cellcolor[HTML]{B7FBFF}0.606 &
  \cellcolor[HTML]{B7FBFF}0.289 & &
  \cellcolor[HTML]{B7FBFF}0.130 &
  \cellcolor[HTML]{B7FBFF}0.096 &
  \cellcolor[HTML]{B7FBFF}0.305 &
  \cellcolor[HTML]{B7FBFF}0.131 &
  \cellcolor[HTML]{B7FBFF}\textbf{0.283} &
  \cellcolor[HTML]{B7FBFF}\textbf{0.108} \\ \cmidrule{1-2}  \cmidrule{4-7} \cmidrule{9-14}
 \cellcolor[HTML]{FFECA8}&
 \cellcolor[HTML]{FFECA8}\cmark & &
  \cellcolor[HTML]{FFECA8}0.555 &
  \cellcolor[HTML]{FFECA8}0.423 &
  \cellcolor[HTML]{FFECA8}0.613 &
  \cellcolor[HTML]{FFECA8}0.244 & &
  \cellcolor[HTML]{FFECA8}0.119 &
  \cellcolor[HTML]{FFECA8}0.079 &
  \cellcolor[HTML]{FFECA8}0.282 &
  \cellcolor[HTML]{FFECA8}0.114 &
  \cellcolor[HTML]{FFECA8}0.261 &
  \cellcolor[HTML]{FFECA8}0.086 \\
\cellcolor[HTML]{FFECA8}\multirow{-2}{*}{REG} &
  \cellcolor[HTML]{FFECA8}\xmark & &
  \cellcolor[HTML]{FFECA8}\textbf{0.573} &
  \cellcolor[HTML]{FFECA8}\textbf{0.481} &
  \cellcolor[HTML]{FFECA8}\textbf{0.667} &
  \cellcolor[HTML]{FFECA8}\textbf{0.319} & &
  \cellcolor[HTML]{FFECA8}0.143 &
  \cellcolor[HTML]{FFECA8}0.093 &
  \cellcolor[HTML]{FFECA8}0.287 &
  \cellcolor[HTML]{FFECA8}0.121 &
  \cellcolor[HTML]{FFECA8}0.282 &
  \cellcolor[HTML]{FFECA8}0.100 \\ \cmidrule{1-2}  \cmidrule{4-7} \cmidrule{9-14}
\end{tabular}%
}
\caption{Results obtained with action labels during inference and \textit{without} antonyms during training. mAP W/M: mean Average Precision with weighted (W) and macro (M) averaging. Coloured rows indicate variants of our method. Bold denotes best result per column.\vspace{-10pt}}
\label{tab:results_no_ant}
\end{table*}

\vspace{-10pt}
\fparagraph{Importance of Context} We now focus on the importance of using textual context for regression comparing \textit{REG - fixed $\delta$} to \textit{REG} in Table~\ref{tab:results_all}. 
Visual features vary differently when the same action is modified by different adverbs, e.g. chopping something coarsely entails a more prominent change than chopping it slowly. With $REG$ we learn such visual differences with targets that change (sometimes considerably) across adverbs
for a given verb. A fixed target for all positive adverbs discards such differences: the model is penalised equally for small/big changes. \textit{REG - fixed $\delta$} does this, and its consistent
worse performance 
validates the idea of using text context as a proxy to learn action changes. 

\vspace{-15pt}
\fparagraph{The Key is in the Optimisation} We now compare \textit{MLP + Act Mod} to our method.
For this \textit{Act Mod} variant we replace the video embedding $f'(x, v)$ with $\theta(f'(x, v))$ in the two losses in~\cite{doughty2020action}, where $\theta$ is the additional MLP. With this configuration, the only difference between our method and Action Modifiers lies in the optimisation. 
In Section~\ref{sec:method} we hypothesised that the optimisation of Action Modifiers might be difficult. Results in Table~\ref{tab:results_all} support our hypothesis. When adding the MLP to \textit{Act Mod,} Acc-A remains virtually identical across all datasets. However mAP W/M drop considerably, especially on HowTo100M Adverbs which is the noisiest dataset. This suggests that increasing the depth of the model makes optimisation more difficult.
We conclude that learning adverbs as pre-defined actions changes rather than as trainable parameters is a better strategy.

\vspace{-15pt}
\fparagraph{Testing without Action Labels} In Table~\ref{tab:results_no_act_gt} we report results obtained without using the action label at test time. 
We test Action Modifiers in the same way as described for our method in Section~\ref{sec:method}.
Experiments are directly comparable between Tables~\ref{tab:results_all} and~\ref{tab:results_no_act_gt} since the tested models are the same. 
We note a remarkable drop in performance 
for all methods. This is expected since the task is now harder. 
In fact 
we need to predict the right adverb from multiple video embeddings obtained querying all verbs. 
Nevertheless, our method still achieves the best global performance in this more challenging setting. 
This shows that our method is more robust and is able to better generalise. 

\vspace{-15pt}
\fparagraph{Learning without Antonyms} We now study the case where adverbs are not paired in antonyms. This is a desirable setting as it 
allows more flexibility in data collection. To train Action Modifiers without antonyms we just sample a random negative adverb as opposed to the antonym for the triplet loss. 
We illustrated in Section~\ref{sec:method} how we train our \textit{REG} variant.
We do not evaluate \textit{REG - fixed $\delta$} here since without antonyms it would be similar to binary classification. 
\textit{CLS} 
by design does not use antonyms. 
Table~\ref{tab:results_no_ant} compares results obtained with and without antonyms. Since we assume we do not have antonym labels here we only evaluate mAP W/M.  
Interestingly, Action Modifiers improves slightly on some datasets when removing antonyms, which shows again that 
using only opposite adverbs for contrastive learning limits the model. 
Discarding antonyms for our regression variant has also a strong positive effective. Since we remove the target $-\delta$ for antonyms and treat opposite adverbs as any other negative class, we are not forcing the model to push antonyms further compared to other negative adverbs. This is beneficial to learn adverbs more generally. 

\vspace{-13pt}
\fparagraph{Regression vs Classification} We now compare \textit{CLS} and \textit{REG} in Tables~\ref{tab:results_all},~\ref{tab:results_no_act_gt} and ~\ref{tab:results_no_ant}. 
When testing without action labels (Table~\ref{tab:results_no_act_gt}) and learning without antonyms (Table~\ref{tab:results_no_ant}) we notice a more frequent improvement of \textit{REG} over \textit{CLS}.
Rather than maximising the margin between the positive adverb and all other negative adverbs as in classification, with regression we try to learn the extent of the change applied by an adverb. This can be a more effective way to learn adverbs 
since the visual change introduced by an adverb varies considerably according to the action. 

\vspace{-3pt}
\section{Conclusion}
\label{sec:conclusion}
\vspace{-2pt}

We proposed to address adverb recognition as a regression task. 
Our method attains new state-of-the-art performance on multiple datasets. Importantly, we 
achieve compelling results when removing two major assumptions: the availability of action labels during testing and the pairing of opposite adverbs as antonyms. 
To address the scarcity of datasets for adverb recognition we introduced AIR. The dataset collects instructional recipe videos where actions are particularly influenced by the modification indicated by adverbs. 
We believe AIR is a valuable resource that can readily foster advance in adverb understanding. 

\vspace{-12pt}
\fparagraph{Limitations} Since we model action changes measuring distances in a text embedding space we rely on a reasonably good text model. If the model does not capture meaningful relationships between verbs and adverbs then we do not have a reliable proxy to learn adverbs. Because we look for action changes directly in the video, 
actions need to exhibit a sufficient visual change when modified by the adverb. 

\vspace{-12pt}
\fparagraph{Future Work} Directions for future work include exploring ways to capture a broader action context via text. For example, the full caption associated with a video can provide additional information (e.g. object nouns) that could be exploited to better learn action modifications. This would require addressing the problem of video-text alignment, as well as the intrinsic noisy nature of long text embeddings.

\vspace{-12pt}
\fparagraph{Potential Societal Impact} We source AIR from HowTo100M dataset, which gathers videos from YouTube. As such, any societal bias introduced in HowTo100M is potentially included in our data and our trained models.

\vspace{-7pt}

\fparagraph{Acknowledgements} Research funded by UKRI through the Edinburgh Laboratory for Integrated Artificial Intelligence (ELIAI) and the Turing Advanced Autonomy project.

{\small
\bibliographystyle{ieee_fullname}
\bibliography{egbib}
}

\pagebreak

\appendix

\section{Adverbs in Recipes - Details}
\label{sec:air_details}

\paragraph{Parsing Captions}
We use SpaCy~\cite{ines_montani_2022_7310816} to parse captions in HowTo100M~\cite{miech19howto100m}. We start filtering captions containing a verb in one of the following tenses/forms: `VB': base verb (e.g. ``take''), `VBP': present tense (e.g. ``take''), `VBZ': present tense 3rd person singular (e.g. ``takes''), `VBG': gerund, present participle (e.g. ``taking''). 
We 
discard verbs in past tenses to avoid parsing adjectives as adverbs (e.g. ``coarsely ground''). 
We then look among the syntactic children of each verb to find adverbs attached to the verb. We manually cluster verbs and adverbs with a similar meaning. We then filter out: i) verbs and adverbs co-occurring less than 100 times; ii) adverbs related to location (e.g. ``diagonally''), feelings (e.g. ``happily''), instants/periods (e.g. ``immediately, continually''), adverbs that are subjective (e.g. ``beautifully'') or too generic (e.g. ``normally''); iii) videos shorter than 5 seconds and longer than 1 minute. 

\vspace{-10pt}

\paragraph{Annotation} After filtering we collected 11,271 video clips, which we annotated via Amazon Mechanical Turk (AMT). For each video we asked annotators to confirm if the action was visible and performed as indicated by the adverb. We also asked additional questions regarding video editing. Specifically, annotators were asked to check if: i) the speed of the video was altered (i.e. slowed-down or sped-up); ii) the video contains jump-cuts (i.e. parts of the action are skipped); iii) the video contains static segments (e.g. still frames with text). We collected these extra annotations for potential future studies. Each video was labelled by 3 annotators. We employed a total of 5 annotators for edge cases where people did not reach a consensus regarding the main questions (action is visible and is performed as indicated by the adverb). We kept videos where the majority confirmed that both the action and the adverb effect are visible, which resulted in 7,003 videos. We showed annotators several examples to illustrate the task. 

\vspace{-10pt}

\begin{figure*}[t]
    \centering
    \includegraphics[width=\columnwidth]{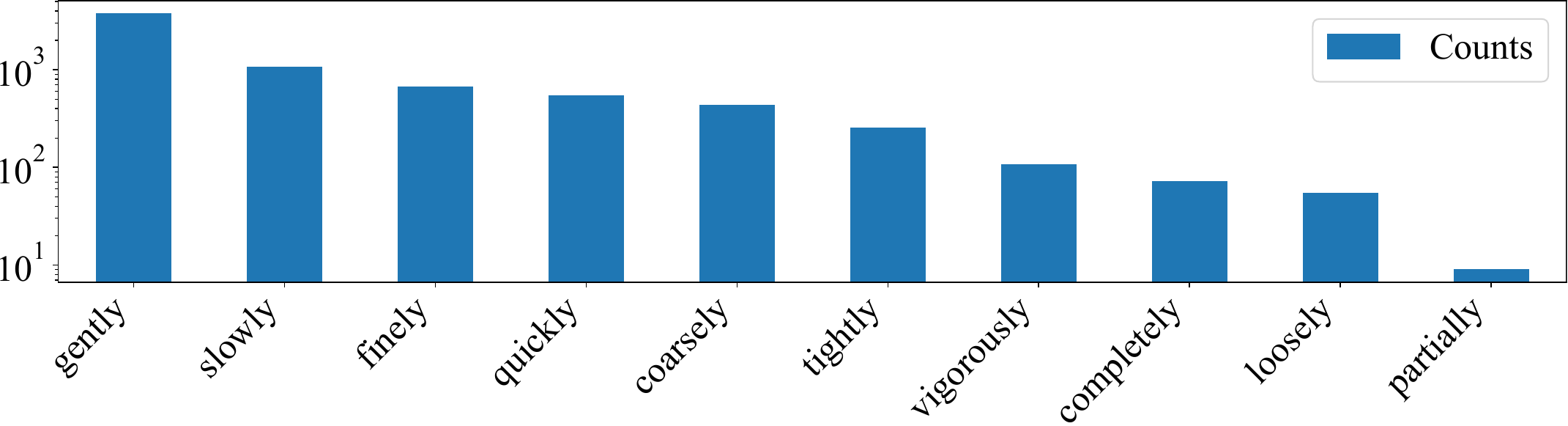}
    \caption{Adverbs in Recipes: adverb distribution (log scale).}
    \label{fig:air_adv_counts}
    \vspace{5pt}
    \centering
    \includegraphics[width=\textwidth]{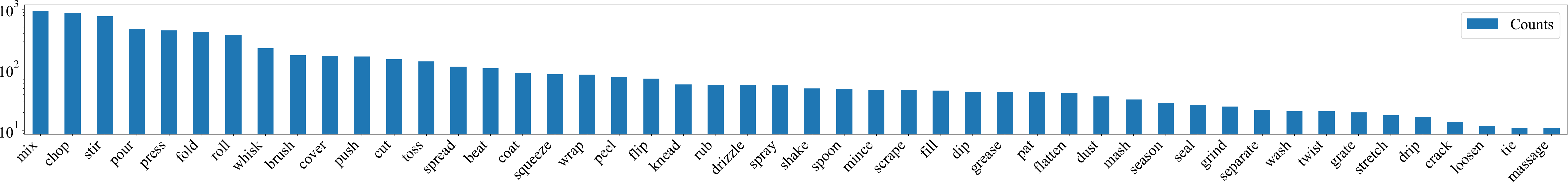}
    \caption{Adverbs in Recipes: verb distribution (log scale).}
    \label{fig:air_verb_counts}
    \vspace{5pt}
    \centering
    \includegraphics[width=\textwidth]{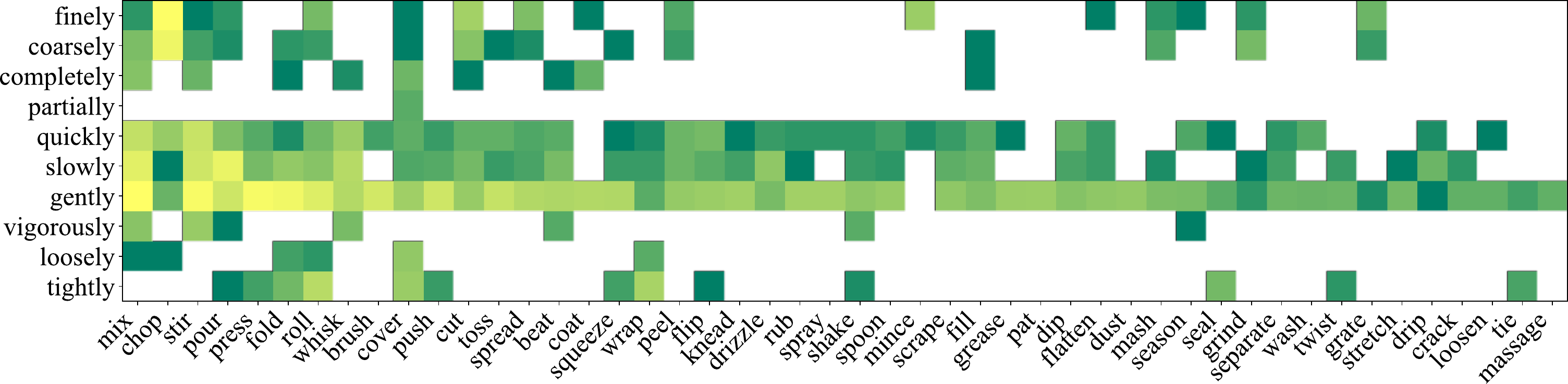}
    \caption{Adverbs in Recipes: verb-adverb co-occurrences. Darker (green)/Brighter (yellow) correspond to less/more frequent pairs. A missing square indicates that the pair does not appear in the dataset.}
    \label{fig:air_co_freq}
\end{figure*}

\paragraph{Verb and Adverb Distributions} We plot the adverb and verb distributions respectively in Figure~\ref{fig:air_adv_counts} and~\ref{fig:air_verb_counts}. Like the existing datasets we reviewed in the paper, 
AIR exhibits a long tail with a heavy class imbalance. 
Figure~\ref{fig:air_co_freq} depicts the co-occurrence matrix of existing (verb, adverb) pairs in the dataset. The matrix is naturally sparse as not all adverbs apply to all verbs. Some 
pairs appear more frequently (e.g. ``chop finely/coarsely'') compared to others (e.g. ``drip gently, mash slowly''). This is expected as some actions and the ways they can be performed are more common than others.

\section{Adverbs in Recipes - Video}
\label{sec:video}
We prepared a video to show a few samples from our new Adverbs in Recipes dataset, which you can watch at \href{https://youtu.be/YPNw35vtyu8}{https://youtu.be/YPNw35vtyu8}. Note how videos are well trimmed and do not contain unrelated content thanks to our better trimming method (see paper for more details). Actions are well visible and importantly are performed as indicated by the adverb, thanks to our manual review.

\begin{table}[t]
\centering
\resizebox{0.9\columnwidth}{!}{%
\begin{tabular}{@{}lcc@{}} \toprule
Parameters              & Our Model & Action Modifiers~\cite{doughty2020action} \\ \midrule
Attention model  (same) & 344,960   & 344,960          \\
Features encoder (MLP)  & 267,786   & -                \\
Adverb parameters       & -         & 2,621,440        \\ \midrule 
Total                   & 612,746   & 2,966,400       \\ \bottomrule
\end{tabular}%
}
\caption{Comparing number of parameters in our model and Action Modifiers~\cite{doughty2020action}, calculated for 10 adverbs. Attention parameters calculated with default settings: 4 heads, input features dimension equal to 1024 and Q, K, V dimensions equal to 512.\vspace{-10pt}}
\label{tab:capacity}
\end{table}

\vspace{-5pt}
\section{Comparing Models Capacity}
\label{sec:capacity}

Table~\ref{tab:capacity} compares the number of parameters in our model and Action Modifiers~\cite{doughty2020action}. We note that our model outperforms Action Modifiers with an order of magnitude fewer parameters ($612,746$ vs $2,966,400$). In particular, our model scales much better according to the number of adverbs. In fact, Action Modifiers learns weights $(E \times E)$ for each adverb. In the experiments $E=512$, thus assuming the number of adverbs $A$ is 10, in Table~\ref{tab:capacity} we have $(E \times E) \times A = (512 \times 512) \times 10 = 2,621,440$. In contrast, only the last layer of our MLP changes according to the number of adverbs $A$. The input to the MLP is a vector of dimension $1024$. We have 3 hidden layers of dimension $512$, whereas the last layer has dimension $A$. Counting both weights and biases, our shallow MLP requires only $267,786$ parameters (assuming $A=10$). The fact that we obtain state-of-the-art results with a much smaller capacity confirms that the key in our better performance lies in a better training strategy. 

\vspace{-5pt}
\section{Results Variance}

\begin{table}[t]
\resizebox{\columnwidth}{!}{%
\begin{tabular}{@{}llcc@{}}
\toprule
Model & mAP W & mAP M & Acc-A \\ \midrule
Act Mod~\cite{doughty2020action} & 0.507 ± 0.003 & 0.248 ± 0.003 & 0.857 ± 0.000 \\
MLP + Act Mod~\cite{doughty2020action} & 0.381 ± 0.001 & 0.142 ± 0.003 & \textbf{0.860} ± 0.000 \\
\rowcolor[HTML]{B7FBFF} 
CLS & 0.605 ± 0.001 & \textbf{0.287} ± 0.001 & 0.841 ± 0.000 \\
\rowcolor[HTML]{FFE0D1} 
REG-fixed $\delta$ & 0.547 ± 0.006 & 0.189 ± 0.003 & 0.836 ± 0.001 \\
\rowcolor[HTML]{FFECA8} 
REG & \textbf{0.611} ± 0.002 & 0.239 ± 0.007 & 0.845 ± 0.001 \\ \bottomrule
\end{tabular}%
}    
\caption{Results variance on AIR. Numbers indicate mean ± std.\vspace{-10pt}}
\label{tab:res-var}
\end{table}

The size of the evaluated datasets is relatively small for deep learning methods. In order to assess the variance of the results we run experiments on AIR two more times, gathering a total of three runs including results from the paper, following the standard setting where models are trained with antonyms and are tested using the action labels. Table~\ref{tab:res-var} reports mean ± standard deviation of the three evaluation metrics. All methods show stable performance. Most importantly, the ranking of the methods and the improvement of our method remains the same.

\end{document}